\documentclass[conference]{IEEEtran}
\IEEEoverridecommandlockouts
\usepackage{float}
\usepackage{graphicx} 
\usepackage{etoolbox}
\makeatletter
\patchcmd{\@makecaption}
  {\scshape}
  {}
  {}
  {}
\makeatother

\usepackage{amsmath,amssymb,amsfonts}
\usepackage{algorithmic}
\usepackage{graphicx}
\usepackage{textcomp}
\usepackage{subfigure}
\usepackage{xcolor}
\usepackage{siunitx} 
\def\BibTeX{{\rm B\kern-.05em{\sc i\kern-.025em b}\kern-.08em
    T\kern-.1667em\lower.7ex\hbox{E}\kern-.125emX}}
\usepackage[backend=biber]{biblatex}
\begin{document}

\title{Fast Point Cloud to Mesh Reconstruction for Deformable Object Tracking\\

}

\author{Elham Amin Mansour$^{1}$, Hehui Zheng$^{1, 2}$, 
Robert K. Katzschmann$^{1}$
\thanks{$^{1}$Soft Robotics Lab, ETH Zürich, Zürich, Switzerland; {\tt\small  \{eaminmans, zhengh, rkk\}@ethz.ch} }%
\thanks{$^{2}$ETH AI Center; Zürich, Switzerland.}%
}


\maketitle

        




\begin{abstract}
    \noindent
    The world around us is full of soft objects we perceive and deform with dexterous hand movements.
  For a robotic hand to control soft objects, it has to acquire online state feedback of the deforming object.
  While RGB-D cameras can collect occluded point clouds at a rate of \SI{30}{\hertz}, this does not represent a continuously trackable object surface.
   Hence, in this work, we developed a method that takes as input a template mesh which is the mesh of an object in its non-deformed state and a deformed point cloud of the same object, and then shapes the template mesh such that it matches the deformed point cloud.
    The reconstruction of meshes from point clouds has long been studied in the field of Computer graphics under 3D reconstruction and 4D reconstruction, however, both lack the speed and generalizability needed for robotics applications. Our model is designed using a point cloud auto-encoder and a Real-NVP architecture.
    Our trained model can perform mesh reconstruction and tracking at a rate of \SI{58}{\hertz} on a template mesh of 3000 vertices and a deformed point cloud of 5000 points and is generalizable to the deformations of six different object categories which are assumed to be made of soft material in our experiments (scissors, hammer, foam brick, cleanser bottle, orange, and dice). The object meshes are taken from the YCB benchmark dataset. An instance of a downstream application can be the control algorithm for a robotic hand that requires online feedback from the state of the manipulated object which would allow online grasp adaptation in a closed-loop manner. Furthermore, the tracking capacity of our method can help in the system identification of deforming objects in a marker-free approach. In future work, we will extend our trained model to generalize beyond six object categories and additionally to real-world deforming point clouds. 

\end{abstract}

   \begin{figure}
    \begin{center}
    \begin{tabular}{c}
     \subfigure[]
        {\includegraphics[width=0.45\textwidth]{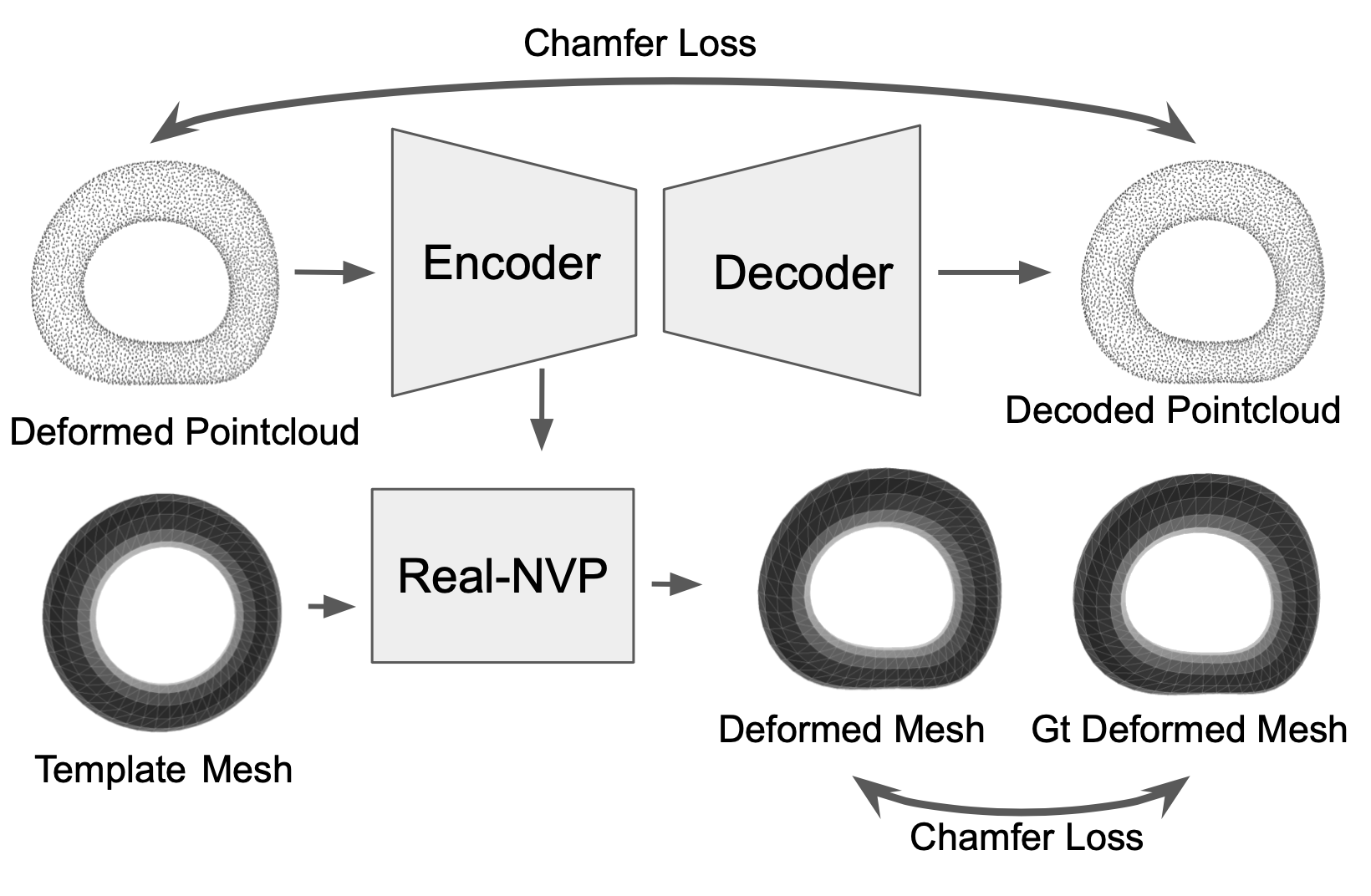}}\\
     \subfigure[]
        {\includegraphics[width=0.42\textwidth]{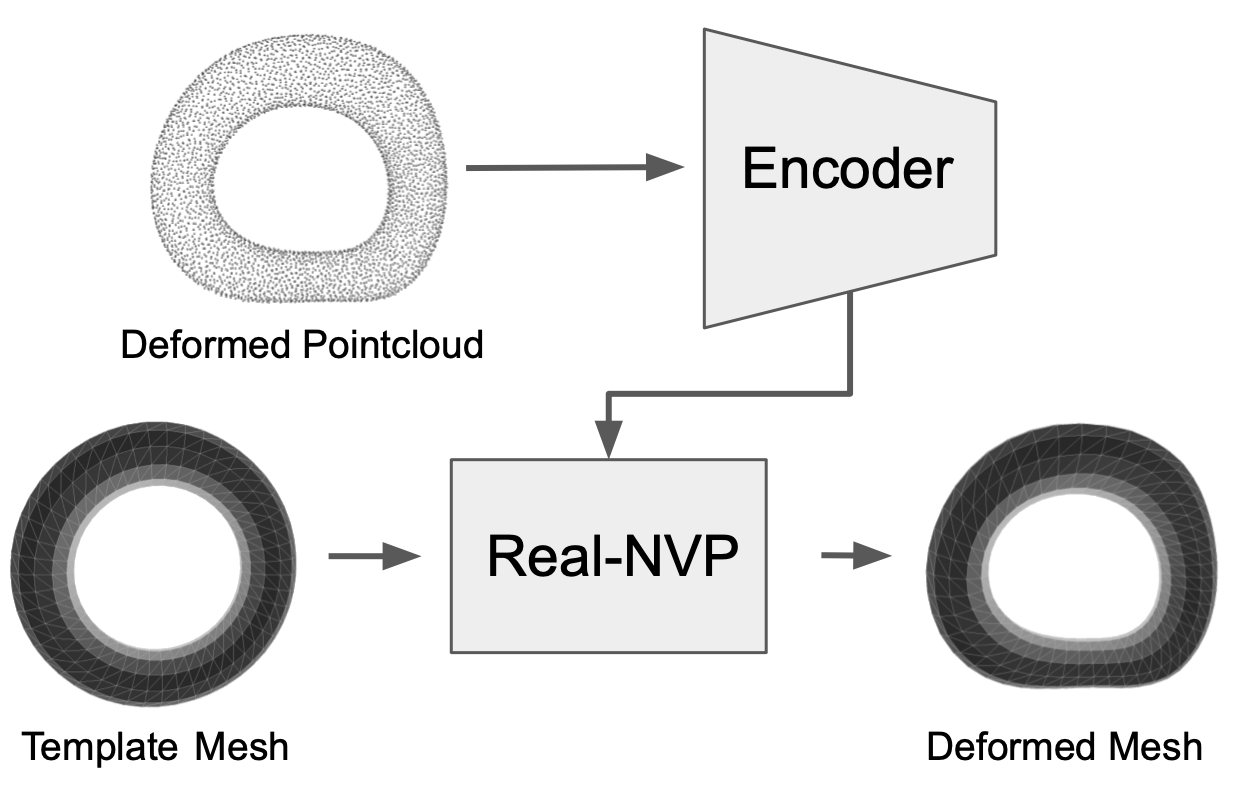}}
    \end{tabular}
        \caption{Our pipeline for the two stages of training (a) and testing (b), respectively, both take as input a deformed point cloud of an object and a mesh of the same object in a non-deformed state (template mesh). It is designed to reconstruct the deformed point cloud into a deformed mesh by deforming the template mesh. (a) Training stage: The auto-encoder, comprised of an encoder and decoder, takes the deformed point cloud as input and learns an encoding through chamfer loss by comparing the decoded/reconstructed deformed point cloud with the groundtruth deformed point cloud.
        Then, the conditional Real-NVP model takes as input the auto-encoder's encoding and the template mesh and learns the coordinates of the deformed mesh using chamfer loss supervised by the ground truth deformed mesh.  
         (b) Inference stage: The encoder encodes the deformed point cloud, and then the conditional Real-NVP model takes the template mesh and the encoding as input and predicts the new coordinate for every vertex in the template mesh. Therefore, in both stages of training and inference, the deformed mesh consists of the template mesh vertices moved around by the Real-NVP, and faces consistent with those of the template mesh.}
        
    \label{fig-method}
    \end{center}
    \end{figure}
\section{Introduction}
The interaction between robots and soft objects is a milestone yet to be hit in the field of robotics. For instance, a robotic hand should be able to quickly learn how an unseen soft object changes shape while it is manipulated and should be able to continuously track the deformations of a soft object to adapt its grasping method accordingly.

The first step towards dexterous manipulation of complex objects is seeing the object. This can be achieved through a model that takes a live stream of point clouds of a deforming object as input and performs an online reconstruction to trackable deforming meshes. The stream of point clouds would be collected via RGB-D cameras positioned at different angles around the object. 

The task of creating meshes from point clouds is a long-studied problem in the computer graphics field. Works in 3D reconstruction generate a mesh for one point cloud~\cite{shapeaspoints,point2mesh,3DN,shapeflow}, while 4D reconstruction works generate meshes for a sequence of deforming point clouds~\cite{LCR4,cadex,NDG,occflow,lpdc}. The main focus of both works is the accurate generation of refined meshes and hence time efficiency is compromised. Furthermore, most of the 4D reconstruction works are based on learning the canonical (or template) mesh, which limits them to training a separate model for each object category with the main focus on humans and animals. In 3D reconstruction, on one hand, implicit algorithms~\cite{shapeaspoints} which use marching cubes are accurate but time inefficient, and on the other hand, explicit models~\cite{point2mesh,3DN,shapeflow} that deform one or more initial template meshes are speed efficient with poor accuracy and topology. Overall, prior works in 3D reconstruction do not apply to the time dimension and hence do not allow for the tracking of the vertex positions of the deformed meshes. This proves to be limiting in robotics applications.


In our work, to find a balance between speed and accuracy, we assume a template mesh can be generated from the first point cloud of the deforming point cloud sequence using a SOTA method, e.g. \cite{shapeaspoints} which operates in the implicit domain and generates meshes of the right topology with a speed of \SI{25}{\hertz}. 
The main focus of our work is generating deformed meshes for the remaining deformed point cloud frames by continuously deforming and tracking the template mesh. In this work, we assume that within a sequence of deforming point clouds of an object, the topology does not change and hence, the template mesh has the same topology as the deformed point cloud of any frame within the sequence. To predict the mesh coordinates of the deformed mesh, we use our model, as illustrated in Fig.~\ref{fig-method}, which consists of an encoder and decoder for extracting an encoding from the deformed point cloud and a Real-NVP~\cite{realnvp} component which deforms the template mesh given the deformed encoding. The resulting pipeline has an inference speed of 0.017 seconds for a template mesh of 3000 vertices and a point cloud of 5000 points on an Omen Desktop(24-core, 3.2GHz Intel Core i9-12900K CPU, 64GB memory, NVIDIA GeForce 3090 GPU with 24GB memory). 

In summary, our main contributions are a trained model:
\begin{itemize}
    \item With a capacity of generating deformable meshes in an online manner of \SI{58}{\hertz} given a template mesh of 3000 vertices and a deformed point cloud of 5000 points.

    \item Generalizable to the deformations of different object categories.
    \item With the capacity to track the vertices.
\end{itemize}

\section{Related Work}

Previous works in the computer graphics field demonstrate point cloud to mesh reconstruction approaches that either work for a single point cloud (3D mesh reconstruction) or a sequence of point clouds (4D mesh reconstruction). In the following, we briefly discuss the most related works and the gaps that need to be addressed.

\subsection{3D Mesh Reconstruction of a Point Cloud}
This category of work reconstructs a separate mesh for each frame of point cloud input, lacking consistent tracking of vertex positions, which can be crucial for robotic applications. These works fall into two categories. In the first category, the vertex positions of a template mesh are modified using a deformation network to best fit the input point cloud that needs to be reconstructed~\cite{point2mesh,3DN,shapeflow,deformAware,nmf}. 

To move around vertices, three different methods are widely used: Neural ordinary differential equations(NODE)~\cite{neuralODE}, Real-NVP~\cite{realnvp}, and MLP. MLP is not homeomorphic and hence does not guarantee manifoldness by the proof in the supplementary material of ShapeFlow~\cite{shapeflow}. While normalizing flows using NODE~\cite{neuralODE} are homeomorphic and guarantee manifoldness, they come at a cost of time inefficiency due to the integrals in their constructions. One pass through the construction takes 0.04s on our hardware. In contrast, a normalizing flow using Real-NVP~\cite{realnvp} guarantees manifoldness at a speed of 0.017s.

Some papers such as 3DN~\cite{3DN}, ShapeFlow~\cite{shapeflow}, and Deformation-Aware 3D Model~\cite{deformAware} use a retrieval network to find a template mesh similar to the input point cloud from a mesh template bank.  The disadvantage of 3DN~\cite{3DN} is that the deforming is done using an MLP which does not guarantee the preservation of the  manifoldness of the mesh. Furthermore, in contrast to our work, these works do not take into account the deformations of one single instance but rather go from one object instance to another object instance. Some other deforming papers such as NMF~\cite{nmf}, Point2Mesh~\cite{point2mesh} and Neural Parts~\cite{neuralparts} start with one or multiple sphere meshes as a template. Point2Mesh~\cite{point2mesh} fits a sphere to one point cloud during inference(which is also its training stage) over three hours which is unfeasible for downstream robotic applications. NMF~\cite{nmf} makes use of flow networks consisting of neural ordinary differential equations to find the flow direction of the sphere template vertices. Therefore, their work, while preserving topology can not reconstruct meshes of a higher genus than zero.
Neural Parts~\cite{neuralparts} deforms five template spheres using Real-NVP~\cite{realnvp}. Starting with multiple template meshes gives the possibility of reconstructing meshes of more complex topology such as a chair with a backseat hole (genus of one). However, this method suffers from disconnectivity between the five resulting component meshes and needs prior knowledge of the number of initial spheres.

The second category of 3D mesh reconstruction papers, models implicit functions such as signed distance functions or an indicator function and then uses marching cubes to reconstruct them. For instance, Shape As Points~\cite{shapeaspoints} uses marching cubes and deep Poisson reconstruction\cite{poissonsurfacerec} and works at a rate of \SI{25}{\hertz} which is the fastest 3D mesh reconstruction in the state of the arts to the best of our knowledge. Additionally, it can reconstruct meshes of arbitrary topology with thin structures. 

Another category of papers combines both the implicit function learning and the flow of the vertices\cite{topologyPreservation}. 
This medical paper\cite{topologyPreservation} is used for the reconstruction of body organs and the point correspondence e.g. livers of different patients. However, due to the template learning of their pipeline, they have a separate pre-trained model for each organ of the human body. Additionally, their inference time is 180.10s which is not feasible for a robotics case of use.

In summary, the fastest model for 3D mesh reconstruction to our knowledge is Shape As Points~\cite{shapeaspoints} with a rate of \SI{25}{\hertz} which can also generate the right mesh topology. However, this method still falls short in speed, particularly in the field of robotic manipulation, and lacks the essential capability of vertex tracking throughout the deformation sequence. 

    \begin{figure}[h!]

    \centering
    \includegraphics[width=8cm]{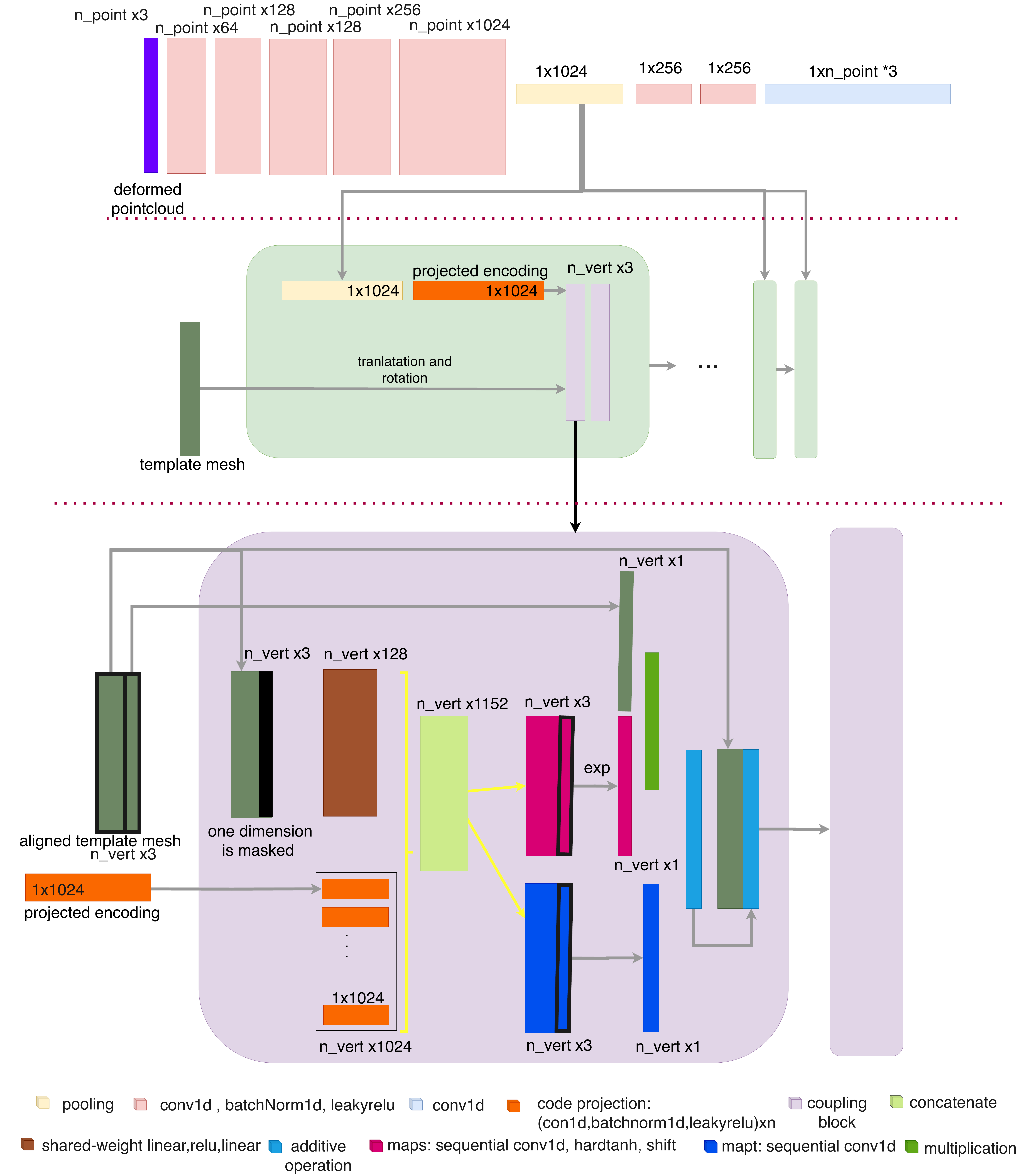}
    \caption{The overall architecture of our method contains an auto-encoder (Top) and a conditional Real-NVP (Middle). The main goal of the model is to deform the mesh of a non-deformed object (template mesh) such that it fits the deformed point cloud of the same object. The auto-encoder takes the coordinates of a deformed point cloud and encodes it into an encoding with one-by-one convolutions and pooling. The Real-NVP which consists of coupling blocks takes the template mesh and the auto-encoder's encoding as input. The architecture within each coupling block (purple block) is shown on the bottom. Within the coupling block, one randomly chosen dimension of the coordinates is masked to zero (randomly chosen for each coupling block) and then the projections of all coordinates to 128 dimensions (brown block) are concatenated to the encoding from the auto-encoder (orange block). This concatenation is shown with a yellow accolade where the encoding is repeated for all coordinates. Subsequently, sequential Conv1d networks called $map_{s}$ (pink block) and $map_{t}$ (dark blue block),  with and without an activation function, are applied to the results. The chosen dimension of the coordinates from the pink block is exponentiated and then multiplied by the corresponding dimension of the template mesh. The result which is the lemon green block is multiplied by the corresponding dimension of the $map_{t}$ network (dark blue block). The result (bright blue block) replaces the chosen dimension of the template mesh while the two other dimensions are kept as they were before. This result is passed on to the next coupling block as the new template and hence the template mesh coordinates are gradually modified to fit the deformed point cloud.}
    \label{arch}
    \end{figure}
\subsection{4D mesh Reconstruction of Point Clouds}
 In the Computer graphics domain, a lot of work has been done in the 4D mesh reconstruction field~\cite{LCR4,cadex,NDG,occflow,lpdc,RFNet-4D}. This consists of firstly, a sequence of point clouds of an object deforming (human or animal in motion, articulated object opening/closing, non-rigid object twisting) being fed to the network. Then, the network predicts a mesh for every point cloud. In some of these papers, the correspondence between meshes (vertex tracking) is also predicted, and in some not~\cite{LCR4}. 
For time-efficiency purposes, some of the works in the 4D field reconstruct only one single mesh of either a canonical state or the first state. This single mesh is referred to as the template mesh. Sequentially like Occupancy Flow~\cite{occflow} or in parallel like CaDeX~\cite{cadex}), the template vertices are mapped to new point coordinates for the remaining point cloud frames. The mesh of each frame is reconstructed using the same faces as the template mesh along with the newly mapped vertex coordinates. 
For instance, the Occupancy Flow~\cite{occflow} paper and the LPDC~\cite{lpdc} paper, use an occupancy field and the multi-resolution isosurface extraction algorithm to reconstruct the mesh of the first point cloud frame and use it as a template mesh. Subsequently, Occupancy Flow~\cite{occflow} uses normalizing flows consisting of neural ordinary differential equations(LPDC~\cite{lpdc} uses MLP) to move around the vertices of the template mesh. 
In CaDeX~\cite{cadex}, the canonical state which is an implicit field is predicted in parallel to vertex tracking. Then, the marching cubes algorithm is used to reconstruct the canonical implicit field as the template mesh. Subsequently, the meshes of other point cloud frames are reconstructed with vertex tracking. This method also uses normalizing flows to move around the vertex coordinates. Its normalizing flow uses Real-NVP~\cite{realnvp} consisting of coupling blocks which is more time efficient than NODE~\cite{neuralODE}. Although CaDeX~\cite{cadex} achieves outstanding results, they have an individual pre-trained model for every specific object category because of their template-learning-based pipeline. For instance, there is a separate pre-trained model for animals, humans, laptops, doors, etc. Similarly,
LPDC~\cite{lpdc} is limited to a pre-trained model for the motion of only humans. Furthermore, these works are not fast enough for the field of robotics: CaDeX~\cite{cadex} takes 0.365s for its rest frames on our machine and 0.68s according to their own recordings. Occupancy Flow~\cite{occflow} takes 0.212s for its rest frames on our machine. The rest frame speed does not take into account the template reconstruction that is done only in the beginning.

The mainstream datasets used in these works Occupancy Flow~\cite{occflow}, CaDeX~\cite{cadex}, LPDC~\cite{lpdc} are deforming human identities from DFAUST~\cite{dfaust}, articulated objects from Shape2Motion~\cite{shape2motion}, humans and animals from DeformingThings4D~\cite{DeformingThings4D} and warped objects using the Occupancy flow\cite{occflow} deformable data generation code.

  \begin{figure*}[h!]
    \centering
    \includegraphics[width=1.\textwidth]{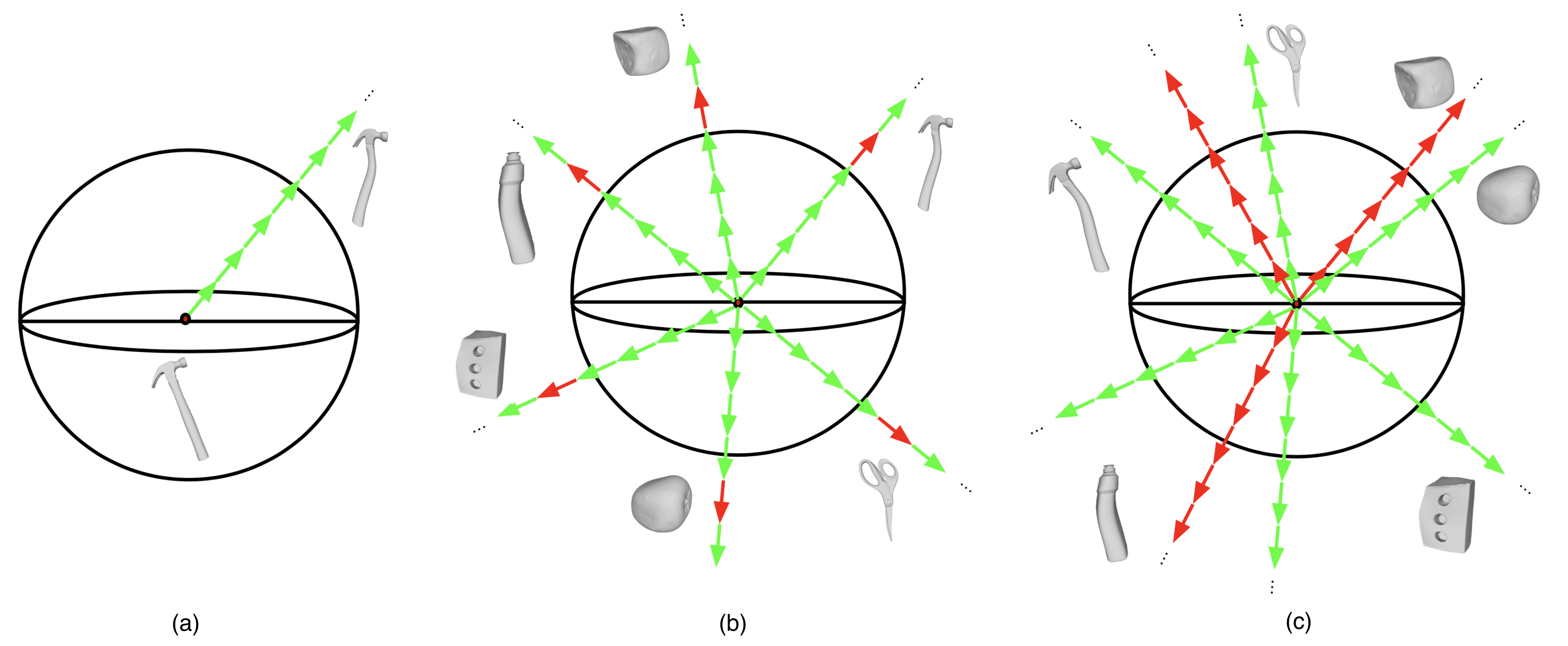}

        \caption{The generation of our different datasets for training and evaluating the model : (a) An arrow in a unique direction represents a unique warping field generated by the deformation code of the Occflow~\cite{occflow} authors. If a warping field is repeatedly applied to a mesh, then a trajectory is created. (b) Dataset B: The red arrows correspond to unseen deformations and the green arrows correspond to seen deformations during training. One trajectory containing 51 deformations is simulated for each of the six instances from the YCB 
        benchmark dataset~\cite{ycb} where steps divisible by five are unseen and the steps indivisible by five are seen. (c) Dataset D: The red arrows correspond to unseen deformations and the green arrows correspond to seen deformations during training. 1000 trajectories are simulated for each of the six YCB instances where 800 of the trajectories are seen and 200 trajectories are unseen. }
        
    \label{exper1}
    \end{figure*}
\section{Methodology}
The overall scheme of our method can be seen in Fig.~\ref{fig-method}. Our model takes as input a mesh of an object in a non-deformed state (template mesh) and a point cloud of the same object in a deformed state. The model learns to deform the template mesh to fit the deformed point cloud. The main components are an auto-encoder and a conditional Real-NVP. The auto-encoder is responsible for the encoding of the deformed point cloud, while the Real-NVP is responsible for the deformation of the template mesh coordinates conditioned on the auto-encoder encoding. Our model can be trained on objects of different categories which also means that during inference, our trained model can be applied to unseen deformations of different object categories as long as the object categories have been seen during training. 
\subsection{Auto-encoder}
During training, an encoder and decoder (auto-encoder model) are firstly pre-trained on the reconstruction of sampled point clouds from our training data using chamfer distance loss reconstruction as can be seen in Fig.~\ref{fig-method}(a). Henceforth, we refer to this loss as the chamfer distance reconstruction loss ($\mathcal{L}_{CDR}$). The encoder learns how to encode vital features of a deformed point cloud such that it can be decoded into a reconstructed point cloud. Subsequently, the Real-NVP is trained with the pre-trained encoder on the same training data. We do ablation studies in Sec.~\ref{dissc} to investigate whether the auto-encoder needs to be frozen while we train the Real-NVP. The pre-training is done in order to ensure that the encoding contains the important features of the point cloud. The final employed auto-encoder model is from \cite{2018auto}. In order to find a suitable architecture, the auto-pointcloud architecture~\cite{2018auto} is compared with FoldingNet architecture~\cite{foldingnet} in Sec.~\ref{dissc}.
\subsection{Real-NVP}
Similarly to prior works~\cite{neuralparts,cadex}, we use Real-NVP in a conditional way. The conditional Real-NVP learns how to deform the template mesh vertices conditioned on the encoding of the deformed point cloud such that it fits the deformed point cloud. Real-NVP consists of several conditional coupling blocks. Each coupling block is a continuous bijective function which makes Real-NVP homeomorphic. Homeomorphisms are continuous bijective maps between two topological spaces that conserve topological properties. The homeomorphic property of the Real-NVP guarantees a stable deformation of the template mesh without causing any holes in the deformed mesh. The coupling block operation which is a conditional affine transformation is explained in Eq.~\ref{eq1}. For each coupling block, one random dimension of the coordinates is chosen and is redefined based on all of the three dimensions of the coordinates and the auto-encoder encoding (which is marked as $enc$ in Eq.~\ref{eq1}) while the two unchosen dimensions of the coordinates remain unchanged. In Eq.~\ref{eq1}, the randomly chosen coordinate dimension is $z$. $map_{s}$ and $map_{t}$ are sequential one-by-one convolutions with and without nonlinearity respectively. The coupling block is applied several times and gradually moves around the template mesh coordinates. The deforming mesh results from the deformed coordinates and shares the same face topology as the template mesh. The learning is done using the chamfer distance loss between the final deformed mesh output from the Real-NVP and the groundtruth deformed mesh as illustrated in Fig.~\ref{fig-method}(a). Henceforth, we refer to this loss as the chamfer distance deformation loss($\mathcal{L}_{CDD}$).

\begin{small}
\begin{equation} \label{eq1}
\left[x^{\prime}, y^{\prime}, z^{\prime}\right]=\left[x, y, z \exp \left(map_s(x, y \mid enc)\right)+map_t(x, y \mid enc)\right]
\end{equation}
\end{small}

\subsection{Network Architecture}
In Fig.~\ref{arch}, the in-depth architecture of our method is presented. In Fig.~\ref{arch}, the auto-encoder is found on the top and the Real-NVP in the middle. The architecture within the coupling block is shown in Fig.~\ref{arch}bottom. The Real-NVP’s main elements are coupling blocks which are homeomorphic (bijective and continuous). Homeomorphisms preserve properties linked to the topology which makes this mapping suitable for stable deformation. Furthermore, a run through the entire models takes about 0.017s for a mesh of 3000 vertices making it all the more suitable for robotic applications. The auto-encoder's main components are one-by-one convolutional layers, batch normalization, rectification non-linearities (ReLU), and max pooling operations. The convolutions help in the extraction of the point cloud features and the pooling maintains the permutation in-variance property. The auto-encoder encodes the deformed point cloud into an encoding and then decodes it back into a point cloud which is compared to the deformed point cloud so that the encoding can be better learned during the pre-training of the auto-encoder. The Real-NVP takes as input the template mesh coordinates and the auto-encoder's encoding. Within each coupling block (purple blocks in Fig.~\ref{arch}), one random dimension of the coordinates is masked to zero. Then the changed coordinates are concatenated to the encoding from the auto-encoder. This concatenation is shown with a yellow accolade in Fig.~\ref{arch}right where the encoding is repeated for each coordinate. Subsequently, sequential one-by-one convolutional networks called $map_{s}$ (pink block) and $map_{t}$ (dark blue block),  with and without an activation function, respectively, are applied to the results. The chosen dimension of the coordinates from the pink block are exponentiated and then multiplied by the corresponding dimension of the template mesh. The result which is the lemon green block is multiplied by the corresponding dimension of the $map_{t}$ network (dark blue block). The result (bright blue block) replaces the chosen dimension of the template mesh while the two other dimensions are kept as they were before. This result is used as the new template for the next coupling block and hence the template mesh gradually deforms into the deformed mesh.


    

\subsection{Inference}
During test time, as illustrated in Fig.~\ref{fig-method}(b), the encoder extracts the encoding from the deformed point cloud and afterwards the Real-NVP deforms the given template mesh using the deformed encoding. The trained model can be applied to unseen deformations of objects it has seen during training. The model can be generalized to different object categories during training. Our final model is trained on the deformations of six different object categories as discussed in Sec.~\ref{gen}. 

\subsection{Dataset}
For the training and testing datasets, the simulation of object deformation is automated using the code from \cite{occflow}.
The code applies random displacement fields to a 3x3x3 grid using a gaussian random variable and then obtains a continuous displacement field by interpolating between the grid points using RBF~\cite{rbf}. The random warping field can then be applied as many times as possible. This consecutive application of the same warping field creates a deformation trajectory containing steps as can be seen in Fig.~\ref{exper1}(a).

 \section{Experiments}\label{res}

\subsection{Dataset Generation}
For the preliminary experiments, we have a small dataset containing deformations of a car and a donut mesh. The donut has 576 vertices and the car contains 15018 vertices. For each mesh, 50 deformed meshes were generated. This is used in the sections \ref{super} and \ref{encoder}. \\ 
Additionally, in order to be able to demonstrate the manipulation capacity, six watertight mesh instances were taken from the ycb\cite{ycb} dataset and four different datasets as explained below were generated using the deformation code. The six watertight meshes include a pair of scissors, a hammer, an orange, a dice, a foam brick, and a bleach-cleanser bottle. Dataset B and D are briefly visualized in Fig. \ref{exper1}(b)(c).
\begin{itemize}
    \item Dataset A: One deformation trajectory containing 50 deformation steps for scissors
    \begin{itemize}
        \item Train set: all deformation steps not divisible by five
        \item Test set: all deformation steps divisible by five
    \end{itemize}
        \item Dataset B: One deformation trajectory containing 50 deformation steps for all six ycb instances
    \begin{itemize}
        \item Train set: all deformation steps not divisible by five
        \item Test set: all deformation steps divisible by five
    \end{itemize}
    \item Dataset C: 1000 deformation trajectories each containing 21 deformation steps for scissors
    \begin{itemize}
        \item Train set: the first 800 trajectories 
        \item Test set: the last 200 trajectories
    \end{itemize}
    \item Dataset D: 1000 deformation trajectories each containing 21 deformation steps for all six ycb instances
    \begin{itemize}
        \item Train set: the first 800 trajectories of each instance
        \item Test set: the last 200 trajectories of each instance
    \end{itemize}

\end{itemize}

\begin{table}[!h]
  \centering
\caption{Comparison of different methods for reconstructing meshes in terms of the essential robotic application properties of generalizability and speed on an Omen desktop computer (24-core, 3.2GHz Intel Core i9-12900K CPU, 64GB memory, NVIDIA GeForce 3090 GPU with 24GB memory).}
\scalebox{.8}{
\begin{tabular}{p{1.5cm}p{0.75cm}p{1.cm}p{4cm}p{1.5cm}}
\hline
\textbf{Work} & \textbf{\textit{Time}} & \textbf{\textit{Method}} & \textbf{\textit{Problem}} & \textbf{\textit{$\#$Categories}}\\
\hline
\textbf{\textit{Ours}} &  0.019s & Real-NVP & Template deformation & 6\\
\hline
\textbf{\textit{CaDeX\cite{cadex}}} &  0.365s & Real-NVP & Subproblem:Template deformation & 1 \\
\hline
 \textbf{\textit{Occflow\cite{occflow}}}&  0.212s & NODE& Subproblem:Template deformation &2\\
\hline
\textbf{\textit{topology\cite{topologyPreservation}}}& 185.27s & NODE& Subproblem:Template deformation &1 \\
\hline
\textbf{\textit{NMF\cite{nmf}}}& 0.14s  & NODE& 3D reconstruction &shapenet\\
\hline
\textbf{\textit{Point2mesh\cite{point2mesh}}}& 3h & MLP & 3D reconstruction&one instance \\
\hline
\end{tabular}}
\label{table1}
\end{table}

    \begin{figure*}[h]
    \centering

    \includegraphics[width=15cm]{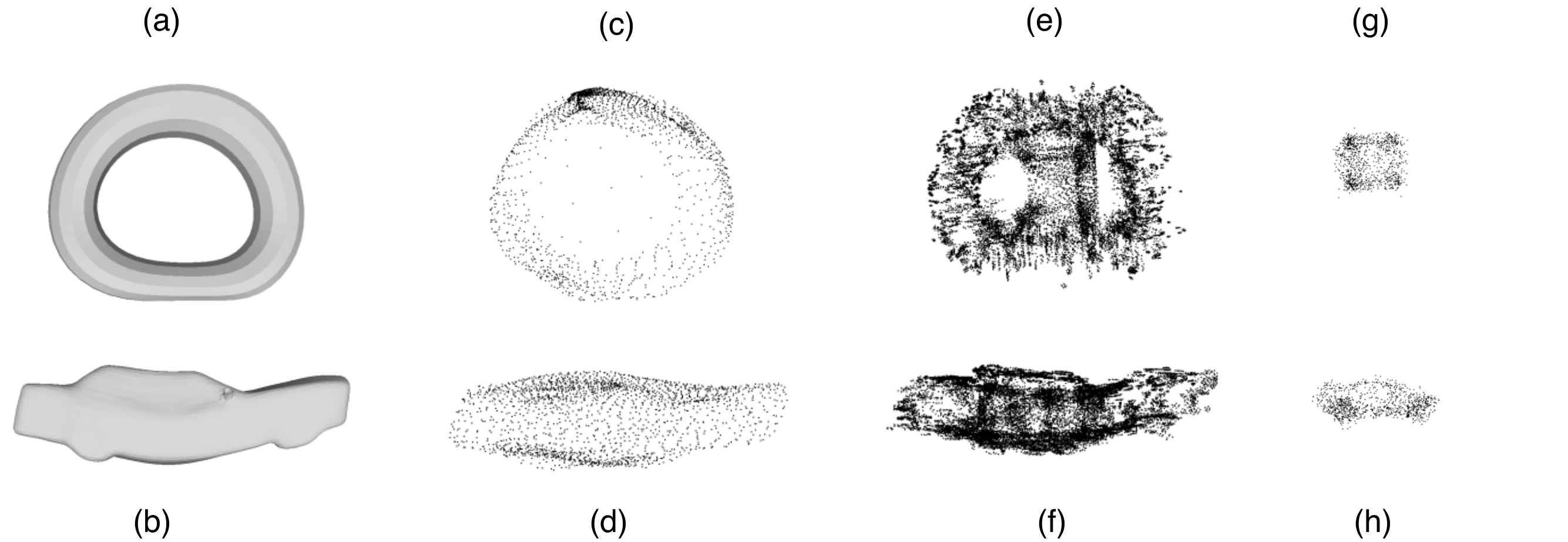}




        \caption{Comparison of different pre-trained encoders trained on 8-10 Shapenet categories(containing no deformations) which contain a car category but not a donut category: (a) Groundtruth mesh of the deformed donut at step 50 (b) Groundtruth mesh of deformed car at step 27 (b) Decoded donut point cloud using FoldingNet~\cite{foldingnet} (d) Decoded car point cloud using FoldingNet~\cite{foldingnet} (e) Decoded donut point cloud using GRNET~\cite{xie2020grnet} (f) Decoded car point cloud using GRNET~\cite{xie2020grnet} (e) Decoded donut using PCN~\cite{yuan2018pcn} (h) Decoded car point cloud using PCN~\cite{yuan2018pcn}.}
        
    \label{pretraineddecoders}
    \end{figure*}

   

\subsection{Discussion}\label{dissc}
 All of the experiments were done on an Omen desktop computer (24-core, 3.2GHz Intel Core i9-12900K CPU, 64GB memory, NVIDIA GeForce 3090 GPU with 24GB memory).
 \subsubsection{Auto-encoder}\label{encoder}
 As the auto-encoder architecture is an important part of this model, several experiments were done in order to explore its capacity and suitable architecture.

 In a first stage, the possibility of using pre-trained encoders from other works was considered. These pre-trained encoders are trained on some of the Shapenet categories (8-10 different categories) where the training data does not contain any deformations. The purpose of this experiment was to see whether such pre-trained models are generalizable to deformations. The reuse of pre-trained encoders would save time and allow our work to focus solely on the training of the conditional Real-NVP. The works that were considered include PCN~\cite{yuan2018pcn}, GRNET~\cite{xie2020grnet}, and FoldingNet~\cite{foldingnet} which are the SOTA of point cloud reconstruction on benchmarks. As can be seen in Fig.~\ref{pretraineddecoders}, qualitatively, none of the pre-trained encoders worked well on the deformed donut which was an unseen category during training. Additionally, PCN~\cite{yuan2018pcn} and GRNET~\cite{xie2020grnet} did not do well on the deformed car which was a seen category. The trained FoldingNet also failed to capture the tires of the car. Thus, in the following experiments, the auto-encoder was retrained on datasets containing deformations. Furthermore, we tried different variations of these auto-encoders by changing the encoding size in order to preserve more features from the deformed point cloud.

  During the training of the encoder, two different architectures were tested, the FoldingNet~\cite{foldingnet} and the Auto-pointcloud~\cite{2018auto}. Additionally, the encoding size of these architectures was also studied. The results in terms of $\mathcal{L}_{CDR}$ and $\mathcal{L}_{CDD}$ are reported in Table~\ref{tableEncoder}. The most indicative metric of the final quality is $\mathcal{L}_{CDD}$ because it compares the final output mesh with the groundtruth mesh. 
  Current literature states that independent one-by-one convolutions maintain permutation invariance of point clouds but do not capture local features. Architectures such as FoldingNet overcome this by applying poolings on local neighborhoods using KNNs. However, in Table~\ref{tableEncoder}, our experiments show a decrease in performance in terms of chamfer distance when using FoldingNet~\cite{foldingnet} in comparison to Auto-pointcloud~\cite{2018auto}. 
   Moreover, the FoldingNet~\cite{foldingnet} architecture and Auto-pointcloud~\cite{2018auto} have better performance in terms of $\mathcal{L}_{CDD}$ when the encoding size is smaller. This is unintuitive as one would expect the deformation to be of better quality if the encoding keeps more information about the deformed point cloud. Our results could be due to slight overfitting in the case of a small encoding size. However, we believe that a larger encoding size would enhance the performance considerably if the model were to be trained on many more object categories. When trained on a very diverse training set, the encoding would have to keep a lot of information in order to be generalizable to many object categories.\\
   In Exp. 5 and 6, reported in Table~\ref{tableEncoder}, we consider whether unfreezing the auto-encoder while training it with the Real-NVP worsens the performance (End2End). It appears that when changing only this End2End factor, the FoldingNet method improves as seen in Table~\ref{tableEncoder} by comparing Exp. 4 and Exp. 6, however, the Auto-pointcloud performance does not change as can be deduced by comparing Exp. 2 and Exp. 5. We continue our experiments with the Auto-pointcloud architecture and encoding size 1024 given that Auto-pointcloud performs better than FoldingNet and the 1024 encoding size allows for more information encoding.

\begin{table}[h]
\caption{Comparison of the performance of different encoders(architecture and bottleneck size are the main studied variants)}

\begin{tabular}{p{0.5cm}p{0.5cm}p{1cm}p{1cm}p{1cm}p{1cm}p{1cm}}
\hline
\textbf{Exp.} & \textbf{\textit{Set}} & \textbf{\textit{Auto-encoder Arch.}} & \textbf{\textit{Embedding Size}} & \textbf{\textit{$\mathcal{L}_{CDR}$}} & \textbf{\textit{$\mathcal{L}_{CDD}$}}& \textbf{\textit{End2end$^{\mathrm{*}}$}}\\
\hline
\textbf{\textit{Exp1}} &  B & Auto2018 & 256 & 0.0010 & 0.0011 & no\\
\hline
\textbf{\textit{Exp2}} &  B & Auto2018 & 1024 & 0.0010 & 0.0012 & no\\
\hline
 \textbf{\textit{Exp3}}&  B & Folding-net & 256 & 0.0011 & 0.0016 & no\\
\hline
 \textbf{\textit{Exp4}}&  B & Folding-net & 1024 & 0.0011 & 0.0017 & no\\
 \hline
 \textbf{\textit{Exp5}}&  B & Auto2018 & 1024 & - & 0.0012 & yes\\
\hline
 \textbf{\textit{Exp6}}&  B &  Folding-net & 1024 & - & 0.0013 & yes\\

\hline

\multicolumn{7}{l}{$^{\mathrm{*}}$The encoder is not frozen while the Real-NVP is trained.}
\end{tabular}
\label{tableEncoder}
\end{table}

\begin{table}[h]
\caption{Generalizability capacity: assessing the generalizability of the model by comparing the performance of a model trained on the deformations of six instances(Datasets B and D) in contrast to one instance(Datasets A and C). The test set of Dataset B contains unseen step deformations and dataset C contains unseen deformation directions. The metric $\mathcal{L}_{CDD}$ indicates the final mesh quality in comparison to the groundtruth deformed mesh.}

\begin{tabular}{p{0.5cm}p{0.5cm}p{0.5cm}p{0.8cm}p{0.5cm}p{0.7cm}p{0.7cm}p{1cm}}
\hline
\textbf{Exp.} & \textbf{\textit{Train set}} & \textbf{\textit{Valid set}} & \textbf{\textit{Auto-encoder Arch.}} & \textbf{\textit{Embed Size}} & \textbf{\textit{$\mathcal{L}_{CDR}$}} & \textbf{\textit{$\mathcal{L}_{CDD}$}}& \textbf{\textit{End2end$^{\mathrm{*}}$}}\\
 \hline
 \textbf{\textit{Exp.7}}&  A & B & Auto2018 & 1024 & 0.0002 & 0.0002 & yes\\
\hline
 \textbf{\textit{Exp.8}}&  B &  B & Auto2018 & 1024 & 0.0007 & 0.0003 & yes\\
\hline
\textbf{\textit{Exp.9}} &  C &C & Auto2018 & 1024 & 0.0002 & 0.0004 & yes \\
\hline
\textbf{\textit{Exp.10}} &  D &C & Auto2018 & 1024 &0.0003 & 0.0005& yes\\
\hline

\multicolumn{8}{l}{$^{\mathrm{*}}$The encoder is not frozen while the Real-NVP is trained.}
\end{tabular}
\label{tablescsixSteps}
\end{table}




\begin{table}[!h]
\centering
\caption{Exp.7 Individual object $\mathcal{L}_{CDD}$}

\begin{tabular}{p{2.5cm}p{1.5cm}}
\hline
\textbf{\textit{Object type}} & \textbf{\textit{$\mathcal{L}_{CDD}$}} \\
\hline
 \textbf{\textit{Scissors}}&  0.0004 \\
 \hline
 \textbf{\textit{Hammer}}&   0.0003\\
\hline
 \textbf{\textit{Dice}}&   0.0016\\
\hline
 \textbf{\textit{Bleach-cleanser}}& 0.0006  \\
\hline
 \textbf{\textit{Brick}}&   0.0015\\
\hline
\end{tabular}
\label{tableE6}
\end{table}

 \subsubsection{Generalizibility}\label{gen}
\begin{figure}[h]
\centering
    \includegraphics[width=8cm]{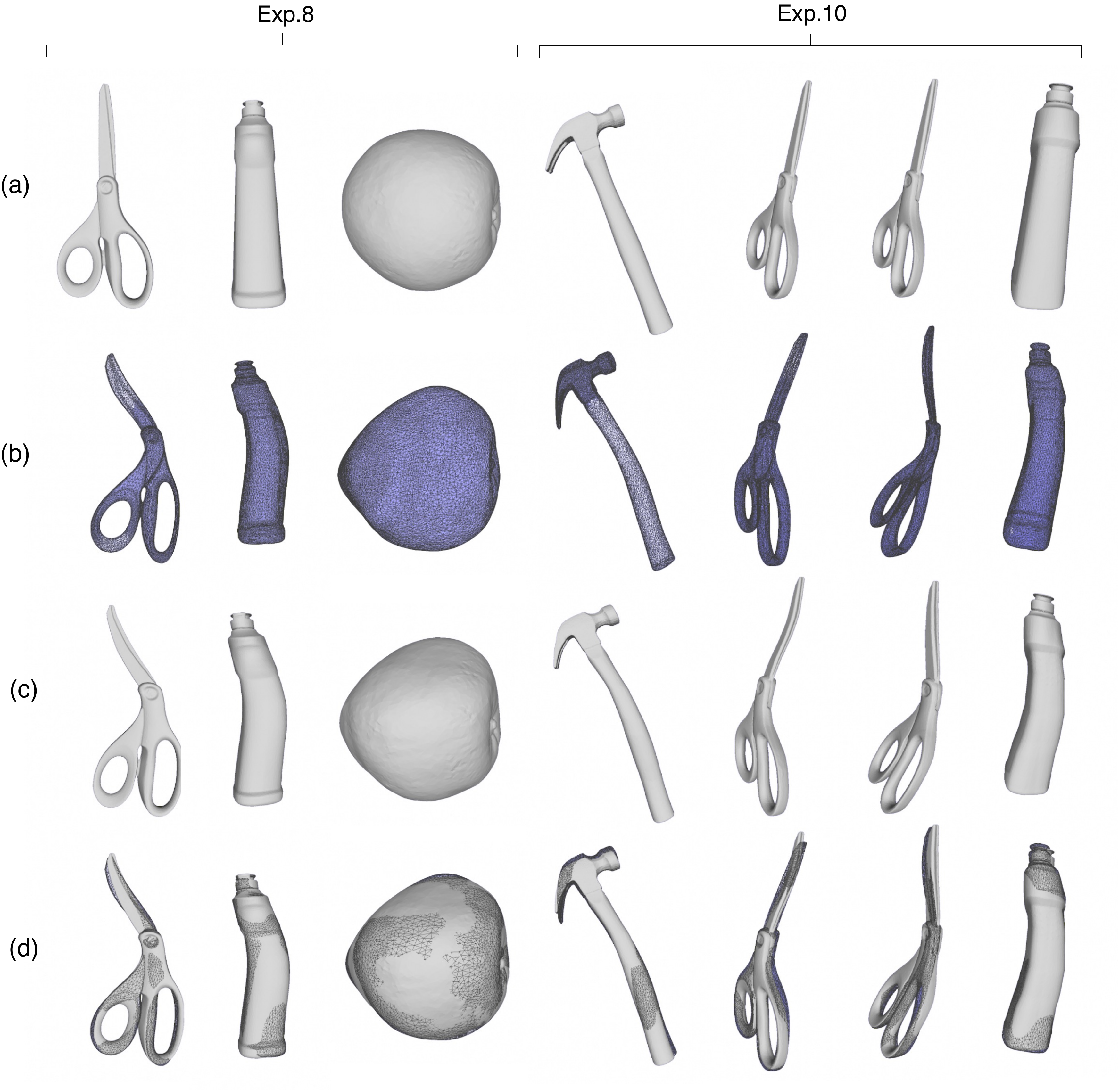}

    \caption{Some visual results of experiments eight and ten(a) Template mesh (b) Groundtruth deformed mesh (c) Predicted mesh (d) Overlay of predicted and groundtruth deformed mesh. Every column corresponds to one deformation.} 
    
\label{allRes}
\end{figure}

\begin{figure}[h]
\centering
    \includegraphics[width=8cm]{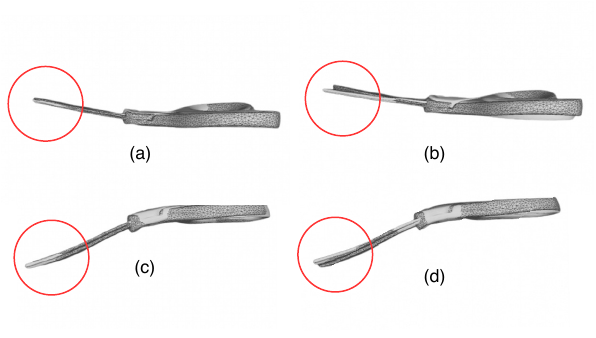}

    \caption{Generalizability of our model: Some visual results from the Exp. 7-10, see Table~\ref{tablescsixSteps}. All of the items visualize the overlay of the predicted and groundtruth deformed mesh.
    (a) Exp. 7 (b) Exp. 8 (c) Exp. 9 (d) Exp. 10, The overlay of (b) is less perfect than (a), and (d) is less perfect than (c) as pointed out in the encircled area. In Exp. 8 and Exp. 10 the model is trained on the deformations of all six YCB instances while in Exp. 7 and Exp. 9 the model is only trained on the deformations of one YCB instance and hence training the model on a larger set of objects worsens the performance on individual objects. (a) and (b) depict the meshes of unseen deformed point clouds from seen deformation trajectories while (c) and (d) depict the meshes of unseen deformed point clouds in unseen deformation trajectories. }
    
\label{generalSc}
\end{figure}

 The generalizability capacity of our method was evaluated on different levels in Table~\ref{tablescsixSteps}. 
Overall, the performance decreases when more generalizability is demanded of the model. The final mesh quality is assessed with the $\mathcal{L}_{CDD}$ metric and therefore is the most indicative when comparing our experiments.  In particular, when comparing the $\mathcal{L}_{CDD}$ of Exp. 7(8) with Exp. 9(10), one can see that evaluating on unseen deformations of seen deformation directions (red arrows in Fig.~\ref{exper1}(b)) has a better performance than unseen deformations of unseen directions (red arrows in Fig.~\ref{exper1}(c)). Furthermore, when the model is trained on more object categories, the evaluation performance on one single category degrades. This can be observed by comparing the $\mathcal{L}_{CDD}$ of Exp. 7(9) with Exp. 8(10) where the training set of B(D) contains deformations of all six items while the training set of A(C) contains only the deformations of one item (scissors). The test set of all four experiments contains only the deformations of scissors. In Fig.~\ref{generalSc}, the visual quality of predicted scissors by models trained on different generalizability conditions is shown. Such generalizability conditions include the number of training object categories and having unseen deformation directions in the test set. Some visual results of Exp. 8 and 10 can be seen in Fig.~\ref{allRes}.

As can be seen in Table~\ref{tableE6}, the numbers depict that the deformation works better for rod-like objects such as the bleach-cleanser, the scissors, and the hammer in contrast to the orange, the dice, and the brick.




 \subsubsection{Comparison of Different Works}
In Table~\ref{table1}, the factors of generalizability and speed of the most relevant works to ours are compared. As discussed before, these two factors are essential for robotics applications. If a method works slower than \SI{30}{\hertz} then its application may not be considered in a robotics context. Furthermore, the generalizability of a method to different object categories is essential for robotics manipulation tasks where any object of an unpredictable category could need manipulation. Therefore, methods that have been designed for one single object category are not suitable for the problem addressed in our work. Some of the works mentioned in Table~\ref{table1}, address our problem only in part. For instance, in the works, OccFlow~\cite{occflow}, CaDeX~\cite{cadex} and Topology\cite{topologyPreservation}, the template mesh is learned in addition to the deformation and hence every trained model works on only one category of objects. For instance, they have individually trained models for each of the categories humans, animals, doors, etc. In contrast, we consider the template mesh learning as a separate problem resolved by SOTA methods such as Shape As Points~\cite{shapeaspoints} and therefore our model remains generalizable to different object categories. Furthermore, despite both our method and CaDeX using Real-NVP~\cite{realnvp}, separating template learning from deformation learning in our approach leads to an increased speed over CaDeX. Moreover, the work, Point2Mesh~\cite{point2mesh}, does not do any vertex tracking in addition to being slow for robotic applications. The NMF~\cite{nmf} is also not time efficient as it works at around \SI{5}{\hertz} which makes it incompatible with robotics settings.

Nevertheless, in order to have a baseline to compare against in terms of quality, we retrained the Occupancy flow~\cite{occflow} model from scratch on our dataset D. In contrast to their training dataset, we have different object categories within the dataset D. The Occupancy flow~\cite{occflow} is the most recent work in the 4D reconstruction SOTA to the best of our knowledge. In Table~\ref{tableCompOcc}, our work is quantitatively compared against Occupancy flow~\cite{occflow}. Our work has a lower chamfer loss. In Fig.~\ref{compOcc1} of the Appendix, our deformed meshes are qualitatively compared against the deformed meshes generated by Occupancy flow~\cite{occflow} for two YCB objects. As illustrated, the bleach cleanser and the scissors are, respectively, missing details and partially reconstructed. Furthermore, the model does not generalize to the 18th-frame deformation of the bleach cleanser. We speculate that because Occupancy flow~\cite{occflow} needs to learn the template mesh, its generalizability capacity to different YCB objects is limited.

 \subsubsection{Adaptive-resolution} \label{super}
Adaptive resolution (a) During training: the template mesh of the car contains 15018 vertices and 30032 faces and hence the deformed mesh contains just as many faces and vertices. (b) During Inference: the template mesh given to the trained model has 27000 vertices and 55172 faces. This shows that the network has the ability to take in templates of different sizes
     \begin{figure}[h]
    \centering
    \begin{tabular}{cc}
     \subfigure[]
        {\includegraphics[width=0.45\textwidth]{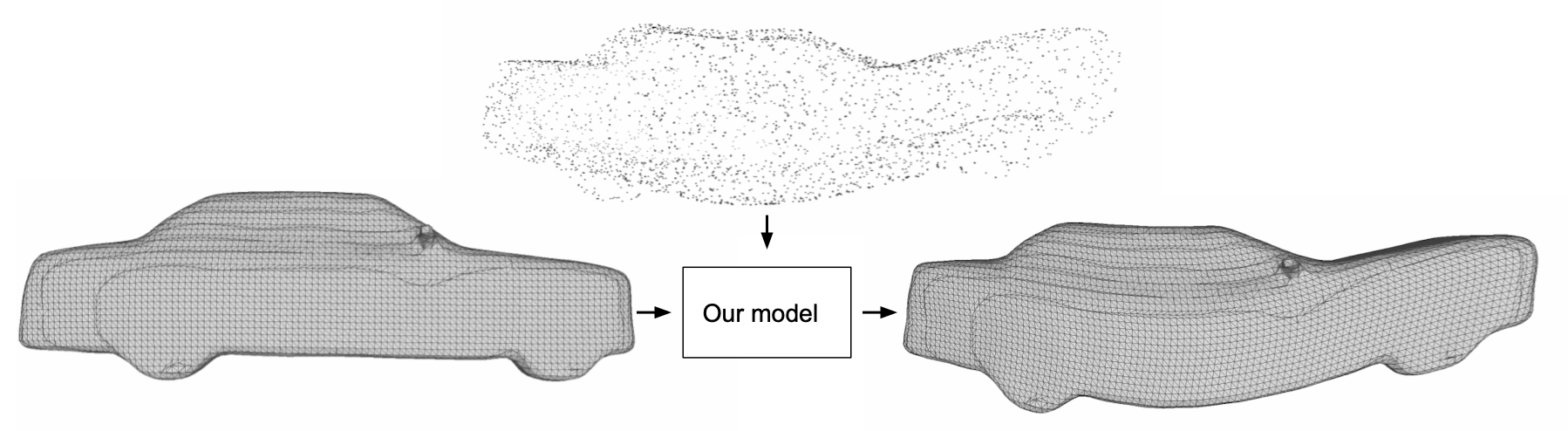}}\\
     \subfigure[]
        {\includegraphics[width=0.45\textwidth]{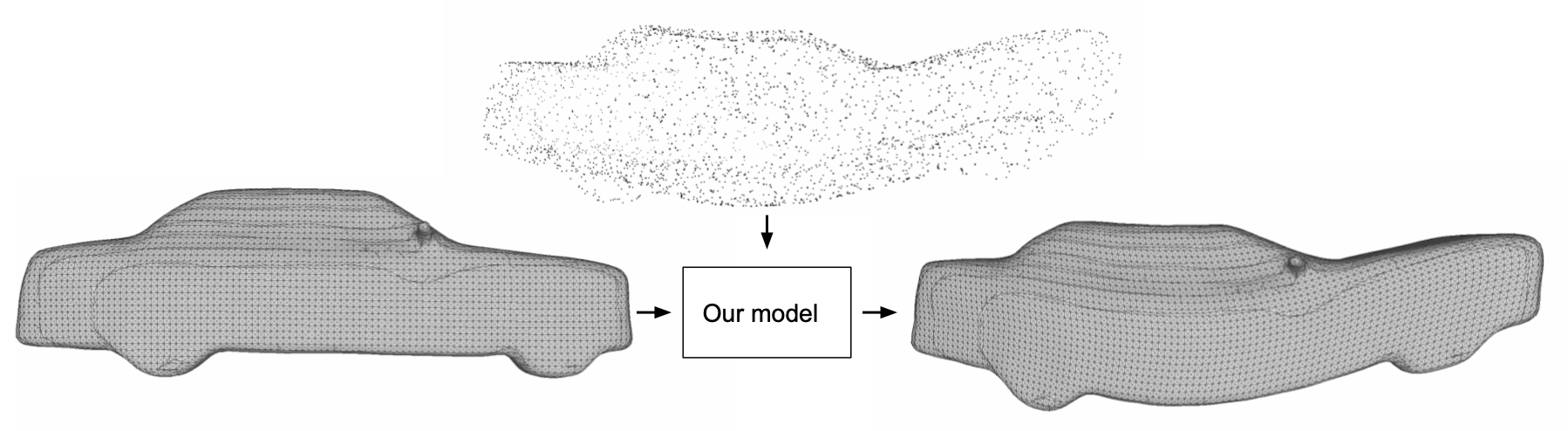}}

    \end{tabular}
        \caption{Adaptive resolution (a) During training: the template mesh of the car contains 15018 vertices and 30032 faces and hence the deformed mesh contains just as many faces and vertices. (b) During Inference: the template mesh given to the trained model has 27000 vertices and 55172 faces. This shows that the network has the ability of taking in vertices of different sizes.}
        
    \label{superfig}
    \end{figure}

\begin{table}[h]
  \centering
\caption{Comparison of $\mathcal{L}_{CDD}$ of our model and the OccFlow model both having been trained on dataset D}
\begin{tabular}{p{2cm}p{2cm}}
\hline
\cline{1-2} 
\textbf{\textit{Method}} & \textbf{\textit{$\mathcal{L}_{CDD}$}}  \\
\hline
ours & 0.0016 \\
\hline
occflow\cite{occflow} & 0.0173\\
\hline
\end{tabular}
\label{tableCompOcc}
\end{table}


    

 \section{Conclusion and Future Work}
 In conclusion, with our method, it is possible to generate meshes from the deforming point clouds(5000 points) of an object and its template mesh(3000 vertices) at a rate of above \SI{50}{\hertz} with simulated data. Additionally, we demonstrate that our model can be generalized to the deformations of at least six objects of different categories and also allows for tracking the vertices of the object. This method has downstream applications such as deformable object manipulation and material learning. In future works, we will focus on firstly generalizing to more object categories which is achievable given higher computational power. In a second step, we will focus on generalizing to real-world point clouds and making the model robust to the occlusions and noises of real-world experiments.
\section{Acknowledgment}
This study was supported by RobotX Research Grant, $\#$RX-03-22, “LPDR - Learning Physics-based Dense Representations for Robotic Manipulation of Soft and Articulated Objects”.

\printbibliography[title={Reference}]

@misc{nmf,
  doi = {10.48550/ARXIV.2007.10973},
  
  url = {https://arxiv.org/abs/2007.10973},
  
  author = {Gupta, Kunal and Chandraker, Manmohan},
  
  title = {Neural Mesh Flow: 3D Manifold Mesh Generation via Diffeomorphic Flows},
  
  publisher = {arXiv},
  
  year = {2020},

  copyright = {arXiv.org perpetual, non-exclusive license}
}

@misc{neuralparts,
  doi = {10.48550/ARXIV.2103.10429},
  
  url = {https://arxiv.org/abs/2103.10429},
  
  author = {Paschalidou, Despoina and Katharopoulos, Angelos and Geiger, Andreas and Fidler, Sanja},
  
  title = {Neural Parts: Learning Expressive 3D Shape Abstractions with Invertible Neural Networks},
  
  publisher = {arXiv},
  
  year = {2021},
  
  copyright = {arXiv.org perpetual, non-exclusive license}
}

@article{point2mesh,
	doi = {10.1145/3386569.3392415},
  
	url = {https://doi.org/10.1145},%2F3386569.3392415}

@misc{3DN,
  doi = {10.48550/ARXIV.1903.03322},
  
  url = {https://arxiv.org/abs/1903.03322},
  
  author = {Wang, Weiyue and Ceylan, Duygu and Mech, Radomir and Neumann, Ulrich},
  
  title = {3DN: 3D Deformation Network},
  
  publisher = {arXiv},
  
  year = {2019},
  
  copyright = {arXiv.org perpetual, non-exclusive license}
}

@misc{shapeflow,
  doi = {10.48550/ARXIV.2006.07982},
  
  url = {https://arxiv.org/abs/2006.07982},
  
  author = {Jiang, Chiyu "Max" and Huang, Jingwei and Tagliasacchi, Andrea and Guibas, Leonidas},
  
  title = {ShapeFlow: Learnable Deformations Among 3D Shapes},
  
  publisher = {arXiv},
  
  year = {2020},
  
  copyright = {arXiv.org perpetual, non-exclusive license}
}

@misc{shapeaspoints,
  doi = {10.48550/ARXIV.2106.03452},
  
  url = {https://arxiv.org/abs/2106.03452},
  
  author = {Peng, Songyou and Jiang, Chiyu "Max" and Liao, Yiyi and Niemeyer, Michael and Pollefeys, Marc and Geiger, Andreas},
  
  title = {Shape As Points: A Differentiable Poisson Solver},
  
  publisher = {arXiv},
  
  year = {2021},
  
  copyright = {arXiv.org perpetual, non-exclusive license}
}

@inproceedings {poissonsurfacerec, 
booktitle = {Symposium on Geometry Processing}, 
editor = {Alla Sheffer and Konrad Polthier}, 
title = {{Poisson Surface Reconstruction}}, 
author = {Kazhdan, Michael and Bolitho, Matthew and Hoppe, Hugues}, 
year = {2006}, 
publisher = {The Eurographics Association}, 
ISBN = {3-905673-24-X}, 
DOI = {10.2312/SGP/SGP06/061-070} 
}

@misc{neuralODE,
  doi = {10.48550/ARXIV.1806.07366},
  
  url = {https://arxiv.org/abs/1806.07366},
  
  author = {Chen, Ricky T. Q. and Rubanova, Yulia and Bettencourt, Jesse and Duvenaud, David},
  
  title = {Neural Ordinary Differential Equations},
  
  publisher = {arXiv},
  
  year = {2018},
  
  copyright = {arXiv.org perpetual, non-exclusive license}
}

@misc{deformAware,
  doi = {10.48550/ARXIV.2004.01228},
  
  url = {https://arxiv.org/abs/2004.01228},
  
  author = {Uy, Mikaela Angelina and Huang, Jingwei and Sung, Minhyuk and Birdal, Tolga and Guibas, Leonidas},
  
  title = {Deformation-Aware 3D Model Embedding and Retrieval},
  
  publisher = {arXiv},
  
  year = {2020},
  
  copyright = {arXiv.org perpetual, non-exclusive license}
}

@misc{cadex,
  doi = {10.48550/ARXIV.2203.16529},
  
  url = {https://arxiv.org/abs/2203.16529},
  
  author = {Lei, Jiahui and Daniilidis, Kostas},
  
  title = {CaDeX: Learning Canonical Deformation Coordinate Space for Dynamic Surface Representation via Neural Homeomorphism},
  
  publisher = {arXiv},
  
  year = {2022},
  
  copyright = {arXiv.org perpetual, non-exclusive license}
}

@article{occflow,
	doi = {10.1109/lra.2022.3151613},
  
	url = {https://doi.org/10.1109},%2Flra.2022.3151613}

@inproceedings{lpdc,
  title={Learning Parallel Dense Correspondence from Spatio-Temporal Descriptors for Efficient and Robust 4D Reconstruction},
  author={Tang, Jiapeng and Xu, Dan and Jia, Kui and Zhang, Lei},
  booktitle={Proceedings of the IEEE/CVF Conference on Computer Vision and Pattern Recognition},
  pages={6022--6031},
  year={2021}
}

@misc{NDG,
  doi = {10.48550/ARXIV.2012.01451},
  
  url = {https://arxiv.org/abs/2012.01451},
  
  author = {Božič, Aljaž and Palafox, Pablo and Zollhöfer, Michael and Thies, Justus and Dai, Angela and Nießner, Matthias},
  
  title = {Neural Deformation Graphs for Globally-consistent Non-rigid Reconstruction},
  
  publisher = {arXiv},
  
  year = {2020},
  
  copyright = {arXiv.org perpetual, non-exclusive license}
}

@misc{LCR4,
  doi = {10.48550/ARXIV.2103.08271},
  
  url = {https://arxiv.org/abs/2103.08271},
  
  author = {Jiang, Boyan and Zhang, Yinda and Wei, Xingkui and Xue, Xiangyang and Fu, Yanwei},
  
  title = {Learning Compositional Representation for 4D Captures with Neural ODE},
  
  publisher = {arXiv},
  
  year = {2021},
  
  copyright = {arXiv.org perpetual, non-exclusive license}
}

@misc{RFNet-4D,
  doi = {10.48550/ARXIV.2203.16482},
  
  url = {https://arxiv.org/abs/2203.16482},
  
  author = {Vu, Tuan-Anh and Nguyen, Duc Thanh and Hua, Binh-Son and Pham, Quang-Hieu and Yeung, Sai-Kit},
  
  title = {RFNet-4D: Joint Object Reconstruction and Flow Estimation from 4D Point Clouds},
  
  publisher = {arXiv},
  
  year = {2022},
  
  copyright = {arXiv.org perpetual, non-exclusive license}
}

@INPROCEEDINGS{dfaust,
  author={Bogo, Federica and Romero, Javier and Pons-Moll, Gerard and Black, Michael J.},
  booktitle={2017 IEEE Conference on Computer Vision and Pattern Recognition (CVPR)}, 
  title={Dynamic FAUST: Registering Human Bodies in Motion}, 
  year={2017},
  volume={},
  number={},
  pages={5573-5582},
  doi={10.1109/CVPR.2017.591}}

@misc{DeformingThings4D,
  doi = {10.48550/ARXIV.2105.01905},
  
  url = {https://arxiv.org/abs/2105.01905},
  
  author = {Li, Yang and Takehara, Hikari and Taketomi, Takafumi and Zheng, Bo and Nießner, Matthias},
  
  title = {4DComplete: Non-Rigid Motion Estimation Beyond the Observable Surface},
  
  publisher = {arXiv},
  
  year = {2021},
  
  copyright = {arXiv.org perpetual, non-exclusive license}
}

@misc{shape2motion,
  doi = {10.48550/ARXIV.1903.03911},
  
  url = {https://arxiv.org/abs/1903.03911},
  
  author = {Wang, Xiaogang and Zhou, Bin and Shi, Yahao and Chen, Xiaowu and Zhao, Qinping and Xu, Kai},
  
  title = {Shape2Motion: Joint Analysis of Motion Parts and Attributes from 3D Shapes},
  
  publisher = {arXiv},
  
  year = {2019},
  
  copyright = {arXiv.org perpetual, non-exclusive license}
}

@article{realnvp,
  title={Density estimation using real nvp},
  author={Dinh, Laurent and Sohl-Dickstein, Jascha and Bengio, Samy},
  journal={arXiv preprint arXiv:1605.08803},
  year={2016}
}

@inproceedings{topologyPreservation,
  title={Topology-preserving shape reconstruction and registration via neural diffeomorphic flow},
  author={Sun, Shanlin and Han, Kun and Kong, Deying and Tang, Hao and Yan, Xiangyi and Xie, Xiaohui},
  booktitle={Proceedings of the IEEE/CVF Conference on Computer Vision and Pattern Recognition},
  pages={20845--20855},
  year={2022}
}

@inproceedings{2018auto,
  title={Learning representations and generative models for 3d point clouds},
  author={Achlioptas, Panos and Diamanti, Olga and Mitliagkas, Ioannis and Guibas, Leonidas},
  booktitle={International conference on machine learning},
  pages={40--49},
  year={2018},
  organization={PMLR}
}

@inproceedings{foldingnet,
  title={Foldingnet: Point cloud auto-encoder via deep grid deformation},
  author={Yang, Yaoqing and Feng, Chen and Shen, Yiru and Tian, Dong},
  booktitle={Proceedings of the IEEE conference on computer vision and pattern recognition},
  pages={206--215},
  year={2018}
}

@inproceedings{yuan2018pcn,
  title={Pcn: Point completion network},
  author={Yuan, Wentao and Khot, Tejas and Held, David and Mertz, Christoph and Hebert, Martial},
  booktitle={2018 international conference on 3D vision (3DV)},
  pages={728--737},
  year={2018},
  organization={IEEE}
}

@inproceedings{xie2020grnet,
  title={Grnet: Gridding residual network for dense point cloud completion},
  author={Xie, Haozhe and Yao, Hongxun and Zhou, Shangchen and Mao, Jiageng and Zhang, Shengping and Sun, Wenxiu},
  booktitle={Computer Vision--ECCV 2020: 16th European Conference, Glasgow, UK, August 23--28, 2020, Proceedings, Part IX},
  pages={365--381},
  year={2020},
  organization={Springer}
}

@article{ycb,
  title={Benchmarking in manipulation research: The ycb object and model set and benchmarking protocols},
  author={Calli, Berk and Walsman, Aaron and Singh, Arjun and Srinivasa, Siddhartha and Abbeel, Pieter and Dollar, Aaron M},
  journal={arXiv preprint arXiv:1502.03143},
  year={2015}
}

@article{rbf,
  title={Universal approximation using radial-basis-function networks},
  author={Park, Jooyoung and Sandberg, Irwin W},
  journal={Neural computation},
  volume={3},
  number={2},
  pages={246--257},
  year={1991},
  publisher={MIT Press}
}
\appendices
\section{Prior method results issues}
In this section, some visualizations are provided in order to give a clearer understanding of the different method performances.
In deforming methods, the mesh is generated by moving around the vertices of the initial mesh in order to fit the target point cloud. For instance, for reconstructing a car point cloud, the deformation network starts off with a car mesh as can be seen in Fig. \ref{fig-template-sim}a or for reconstructing a chair point cloud, the network starts off with a similar chair mesh as can be seen in Fig. \ref{fig-template-sim}b. 
The method Point2mesh\cite{point2mesh} results in non-water tight meshes as can be seen in Fig. \ref{fig-giraffe}.
The NMF\cite{nmf} method, which deforms a sphere, can not reconstruct meshes of a higher genus than zero as can be seen in Fig. \ref{nmf-chair}. The NMF\cite{nmf} is incapable to fit thin structures such as lamps as can be seen in Fig. \ref{nmf-lamp}.
The neural parts\cite{neuralparts} method deforms five template spheres as can be seen in Fig. \ref{parts-human}. The latter gives the possibility of reconstructing meshes of more complex topology as can be seen in Fig. \ref{parts-chair}

Works using implicit models for mesh reconstruction, can reconstruct meshes of arbitrary topology with thin structures as can be seen in Fig. \ref{sap-chair} and \ref{sap-lamp}.

    \begin{figure}[!h]
    \centering
    \begin{tabular}{cc}
     \subfigure[]
        {\includegraphics[width=0.2\textwidth]{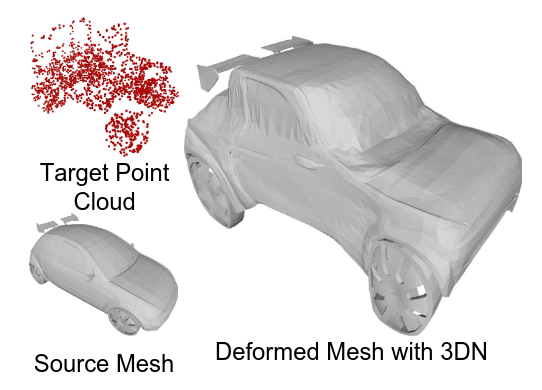}}
     \subfigure[]
        {\includegraphics[width=0.25\textwidth]{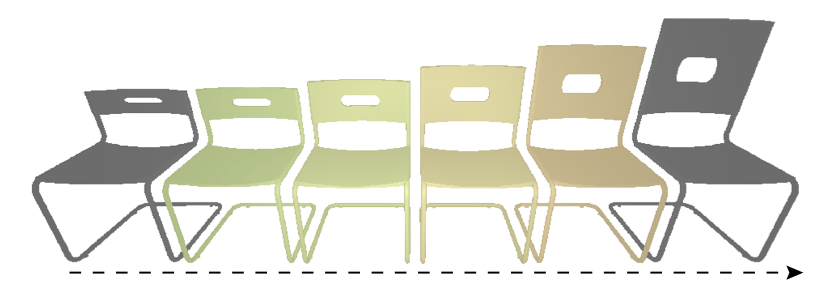}}
    \end{tabular}
        \caption{(a) Image from 3DN\cite{3DN} which shows the result of deforming a source mesh in order to fit a target point cloud (b) Image from the shapeflow\cite{shapeflow} paper that shows the deformations of a chair from a source mesh to a target mesh. In both of these methods, the final mesh is reconstructed by choosing a mesh similar to the input point cloud and moving around slightly its vertices while preserving the mesh faces.}
        
    \label{fig-template-sim}
    \end{figure}

\begin{figure}[!h]
    \centering
    \includegraphics[width=9cm]{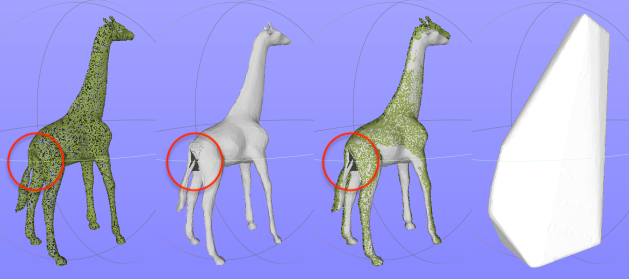}
    \caption{Rerun of the point2mesh code\cite{point2mesh} code: From left to right: the input point cloud, the deformed output mesh, the overlay of the point cloud and the deformed mesh, the template bounding box from which the deformation is started from. The mesh reconstructed with this method is not water-tight as can be seen in the encircled hole.}
    \label{fig-giraffe}
\end{figure}

\begin{figure}[!h]
    \centering
    \includegraphics[width=8cm]{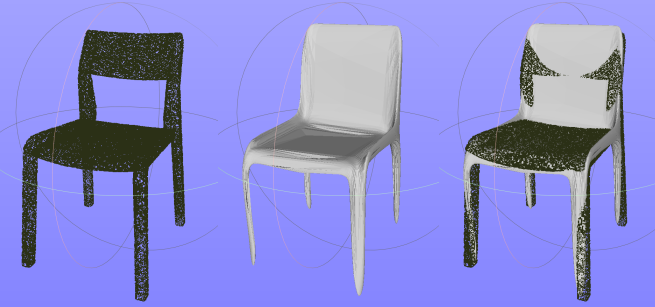}
    \caption{The NMF code\cite{nmf} rerun: from left to right: the input point cloud, the deformed mesh, the overlay of the input point cloud and the deformed mesh. While the groundtruth mesh has one hole in the chair back-rest, the reconstructed mesh does not have any which shows that the model is incapable of capturing the correct topology. Additionally the chair legs in the reconstructed mesh are spiky.}
     \label{nmf-chair}
\end{figure}

\begin{figure}[!h]
    \centering
    \includegraphics[width=8cm]{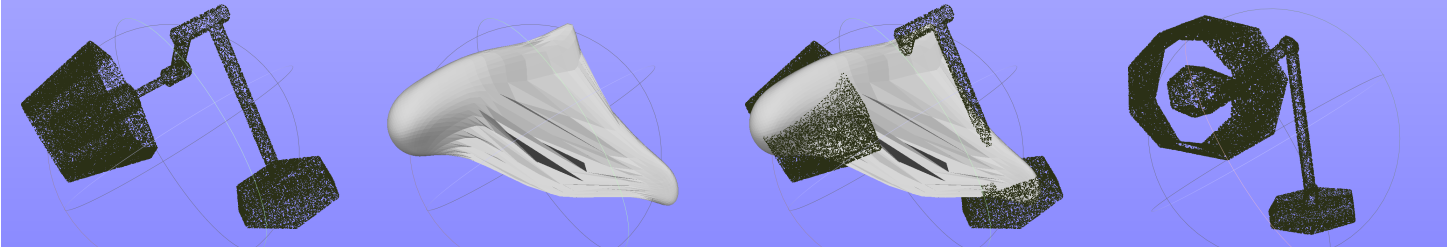}
    \caption{The NMF code\cite{nmf} rerun: from left to right: the input point cloud, the deformed mesh, the overlay of the input point cloud and the deformed mesh, the lamp point cloud input shown from a different angle. The NMF model can not reconstruct the thin structure of the lamp nor the topology of genus one(one hole).}
    \label{nmf-lamp}
\end{figure}

\begin{figure}[!h]
    \centering
    \includegraphics[width=8cm]{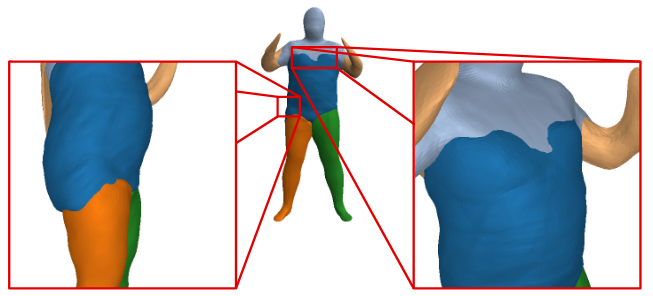}
    \caption{Image from the Neural Parts paper\cite{neuralparts} where each color represents the deformation of a sphere: there are five spheres}
    \label{parts-human}
 \end{figure}


    \begin{figure}[!h]
    \centering
    \includegraphics[width=8cm]{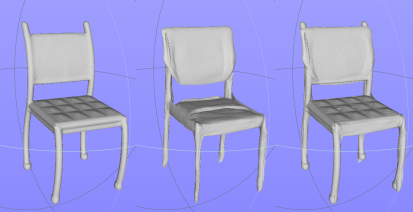}
    \caption{The rerun of the Neural parts code\cite{neuralparts}: from left to right: the groundtruth mesh, the deformed mesh using five spherical template meshes as template and input images of a chair, the overlay of the groundtruth mesh and the output mesh. The final mesh captures the hole inside the chair back-rest however the final mesh is composed of multiple disconnected meshes.}
    \label{parts-chair}
    \end{figure}

    \begin{figure}[!h]
\centering
\includegraphics[width=8cm]{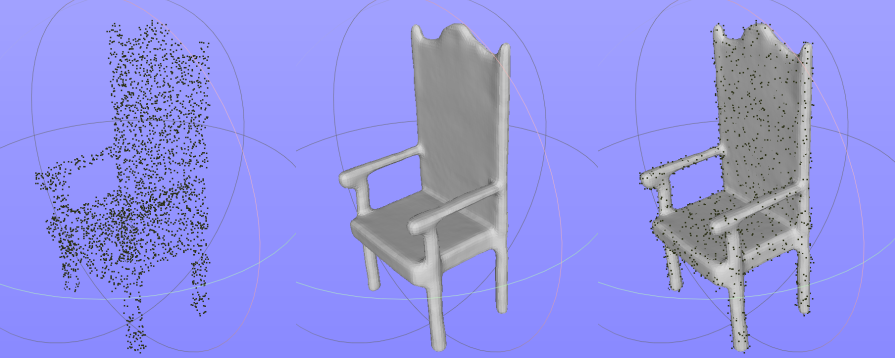}
\caption{The shape as points\cite{shapeaspoints} rerun: from left to right: the input point cloud, the reconstructed mesh of the chair, the overlay of the output mesh and the input point cloud. The topology of the chair which is of genus 2 is well reconstructed because the holes under the chair handles are not filled.}
\label{sap-chair}
\end{figure}
  \begin{figure}[!h]
\centering
\includegraphics[width=8cm]{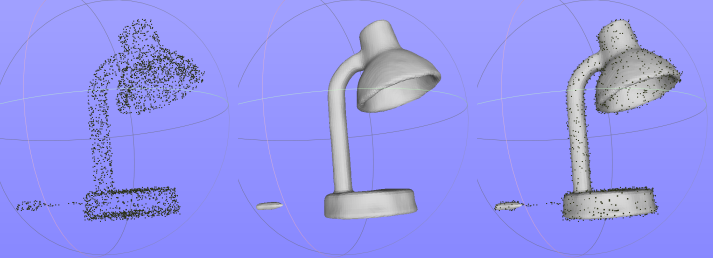}
\caption{The shape as points\cite{shapeaspoints} rerun: from left to right: the input point cloud, the reconstructed mesh of the lamp, the overlay of the output mesh and the input point cloud. The thin structure is well reconstructed. The disconnectivity is due to the sparsity of the input point cloud.}
\label{sap-lamp}
\end{figure}



In Fig. \ref{compOcc1} the results of our method and the results of the method Occupancy flow\cite{occflow} both trained on our dataset D are visualized.

 \begin{figure}[h]
\centering
    \includegraphics[width=5cm]{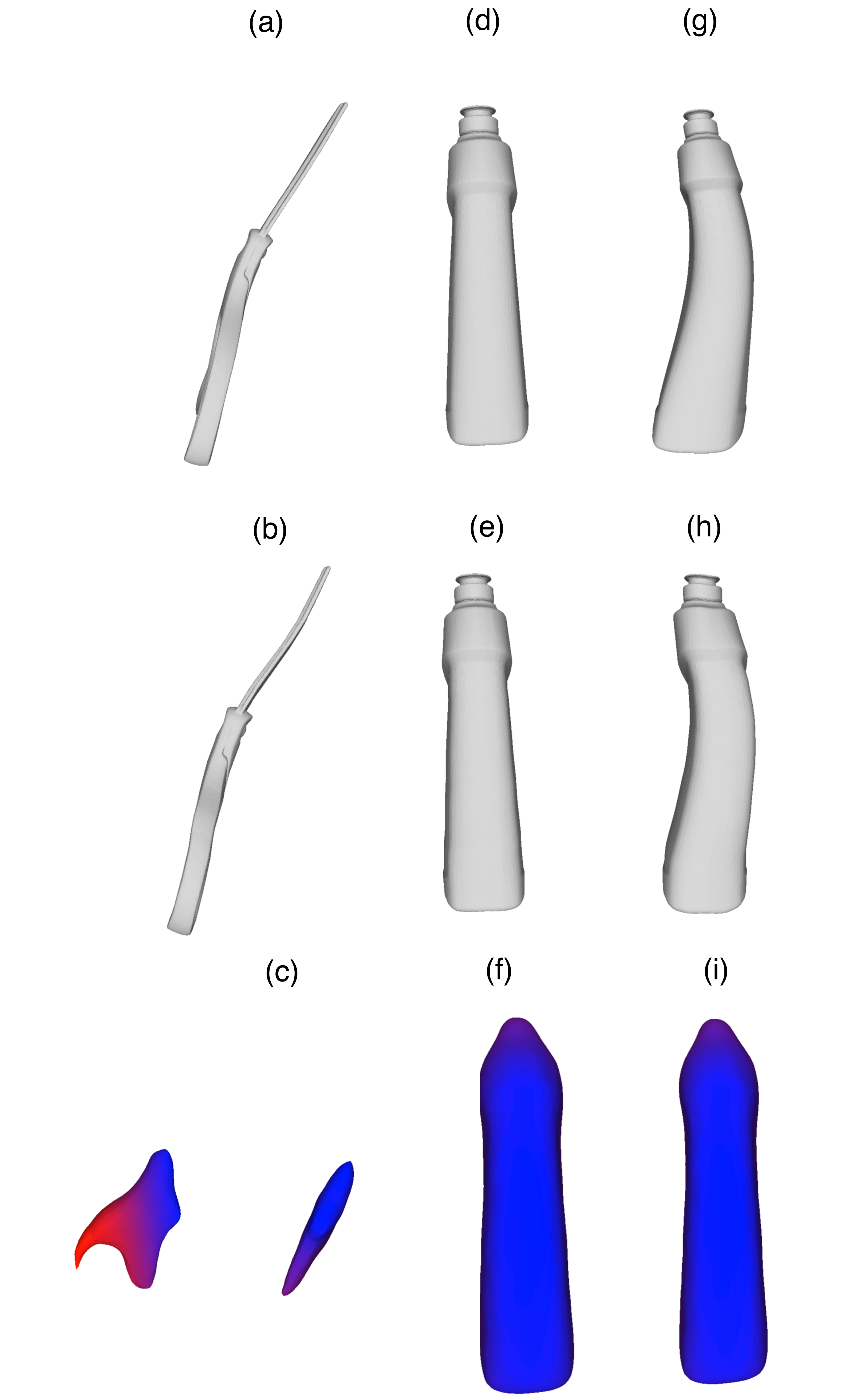}
    \caption{Visual comparison of our method's results and Occupancy flow\cite{occflow} (a) Groundtruth deformed scissor mesh(side view) (b) Our deformed mesh(side view) (c) Occupany flow deformed mesh(side/front view) {d) Groundtruth 1st frame of bleach cleanser (e) Our 1st frame deformed mesh (f) Occupancy flow 1st frame deformed mesh (g) Groundtruth 17th deformed frame of bleach cleanser (h) Our 17th frame deformed mesh  (i) Occupancy flow 17th frame deformed mesh}
    }
    
\label{compOcc1}
\end{figure}

\end{document}